\documentclass{Interspeech2024}
\usepackage{verbatim}
\usepackage{cite}
\usepackage{bbding}
\usepackage{pifont}
\usepackage{hyperref}
\usepackage{makecell}

\interspeechcameraready

\title{Enhancing Voice Wake-Up for Dysarthria: Mandarin Dysarthria Speech Corpus Release and Customized System Design}
\name[affiliation={1}]{Ming}{Gao}
\name[affiliation={1}]{Hang}{Chen}
\name[affiliation={1,*}]{Jun}{Du}
\name[affiliation={2}]{Xin}{Xu}
\name[affiliation={2}]{Hongxiao}{Guo}
\name[affiliation={2}]{Hui}{Bu}
\name[affiliation={3}]{Jianxing}{Yang}
\name[affiliation={4}]{Ming}{Li}
\name[affiliation={5}]{Chin-Hui}{Lee}
\address{
  $^1$\thanks{*Corresponding author}University of Science and Technology of China, China 
  $^2$Beijing AISHELL Technology Co., Ltd., China 
  $^3$Beijing Polytechnic, China 
  $^4$Duke Kunshan University, China 
  $^5$Georgia Institute of Technology, USA}
\email{jundu@ustc.edu.cn}


\begin{document}
\ninept
\maketitle
\vspace{20pt}

\begin{abstract}
    Smart home technology has gained widespread adoption, facilitating effortless control of devices through voice commands. However, individuals with dysarthria, a motor speech disorder, face challenges due to the variability of their speech. This paper addresses the wake-up word spotting (WWS) task for dysarthric individuals, aiming to integrate them into real-world applications. To support this, we release the open-source Mandarin Dysarthria Speech Corpus (MDSC), a dataset designed for dysarthric individuals in home environments. MDSC encompasses information on age, gender, disease types, and intelligibility evaluations. Furthermore, we perform comprehensive experimental analysis on MDSC, highlighting the challenges encountered. We also develop a customized dysarthria WWS system that showcases robustness in handling intelligibility and achieving exceptional performance. MDSC will be released on \url{https://www.aishelltech.com/AISHELL\_6B}.
\end{abstract}

\section{Introduction}
With the rapid advancement of artificial intelligence, voice-controlled applications have become an essential part of our lives. As the initial step towards enabling convenience, voice wake-up has received significant research attention~\cite{ref1,ref2,ref3,ref4,ref5}. However, the rising prevalence of voice wake-up technology poses a potential exclusionary risk for individuals with dysarthria. Dysarthria is a speech disorder characterized by impairments in articulation, fluency, volume, clarity, and pace~\cite{char,dsr}. People with dysarthria due to neurological conditions like cerebral palsy and Parkinson's disease rely heavily on voice-activated technology to meet their daily needs. It is crucial to prioritize inclusivity in designing applications that include dysarthria users. 

In recent decades, the availability of dysarthria corpora~\cite{whitaker,nemours} has catalyzed technological advances and significantly improved the robustness of speech-controlled systems in the face of dysarthria challenges. The most widely used databases are UA-Speech\cite{uaspeech} and Torgo\cite{torgo}. UA-Speech focuses on cerebral palsy patients, capturing speech data using an eight-microphone array and synchronized video recordings. Torgo focuses on cerebral palsy and amyotrophic lateral sclerosis patients, collecting acoustic and articulatory organ motion data through electromagnetic articulography. EasyCall\cite{easycall} incorporates command-like terms, serving as a resource for developing Italian command recognition systems. IDEA\cite{idea} is also an Italian articulatory disorder database.

Recent research has shifted towards collecting speech data in diverse, naturalistic environments, exemplified by the Euphonia\cite{euphonia}, which offers a substantial amount of real-life speech data. Three reported databases for Chinese articulatory disorders: CUDYS\cite{cudys} focuses on acoustic features of pronunciation and prosody. MSDM\cite{msdm} targets subacute stroke patients, capturing audio and facial motion data. However, both databases have small datasets (\textless 10 hours), limiting their use for speech recognition models. CDSD\cite{cdsd} records audio and video data for Chinese dysarthria speech recognition. To our knowledge, there is no dedicated Mandarin dysarthria speech database for wake-up spotting tasks.

In light of this, this paper proposes the first Mandarin dysarthria wake-up word corpus called Mandarin Dysarthria Speech Corpus (MDSC) and a customized WWS system to make voice-activated technologies more accessible for individuals with dysarthria. The main contributions of our study can be outlined as follows:
\begin{itemize}
\item releasing the MDSC featuring 9.4 hours of recordings from 21 dysarthric speakers and 7.6 hours of corresponding control recordings from speakers with standard speech patterns.
\item conducting comprehensive experimental analysis on the MDSC, revealing two major challenges: significant in-domain variance and limited data volume.
\item proposing a customized system robust to intelligibility for dysarthria WWS tasks, demonstrating outstanding performance.
\end{itemize}

\section{MDSC}
\subsection{Statistics}
\begin{table}[!t]
  \caption{Statistical information of every subset in MDSC}
  \label{part}
  \centering
  \setlength{\tabcolsep}{5pt}{
  \begin{tabular}{c|ccc|ccc}
    \toprule
    \textbf{Subset} & \multicolumn{3}{c|}{\textbf{Control}} & \multicolumn{3}{c}{\textbf{Dysarthria}} \\
    & \textbf{Train} & \textbf{Dev} & \textbf{Test} & \textbf{Train} & \textbf{Dev} & \textbf{Test} \\
    \midrule
    \textbf{Duration (h)} & 6.4 & 0.6 & 0.6 & 6.1 & 0.9 & 2.4 \\
    \textbf{Speaker} & 21 & 2 & 2& 13 & 2 & 6 \\
    \bottomrule
  \end{tabular}}
\end{table}

MDSC includes $18,630$ recordings totaling $17$ hours, of which $10,125$ are from non-dysarthric recordings (Control) totaling $7.6$ hours, and $8,505$ are from dysarthric recordings (Dysarthria) totaling 9.4 hours. We record utterances from $21$ dysarthric ($12$ females, $9$ males) and $25$ non-dysarthric ($13$ females, $12$ males) speakers. The participants with dysarthric speakers have the following characteristics:
\begin{itemize}
\item Native Mandarin speakers;
\item Broad age distribution (from $18$ to $48$) and gender balance;
\item Diverse etiologies contribute to dysarthria, including cerebral palsy and hepatolenticular degeneration;
\end{itemize}

Table~\ref{part} delineates the distribution of Control and Dysarthria across training, development, and test sets, providing specifics on the duration and speakers' count. There is no overlap in speakers among the data in each subset. 

Within the Dysarthria test set (D-test), six individuals are assigned names D1-D6 in descending order of intelligibility, ensuring a balanced representation of genders. The corresponding test data for each individual is specifically labeled as D1-test to D6-test. D1 demonstrates the highest level of intelligibility, exhibiting pronunciation nearly indistinguishable from individuals without dysarthria. Conversely, D6 shows the lowest level of intelligibility, characterized by unclear articulation, effortful pronunciation, slower speech rate, and issues with insufficient breath support. For each of these six individuals, we reserve $3$ minutes of enrollment utterances (D1-enroll to D6-enroll) in advance, not included in any subset.

\subsection{Collection}
The recordings consist of 10 wake-up words repeated five times at varying speeds. MDSC also includes 355 non-wake-up words, encompassing fixed command words, free command words, household instructions, and other phrases. The single-person text list has 295 non-repeated sentences.

We transcribe the recordings using our self-developed mobile application. The recordings, sampled at 16kHz, take place in a quiet indoor environment, with the participants positioned approximately 20cm (followed \cite{aishell,aishell2}) away from the microphone.

\subsection{Intelligibility Evaluation}
We employ a comprehensive evaluation framework that combines subjective and objective criteria to quantify the intelligibility of recordings from individuals with dysarthria.

The first criterion is annotation accuracy, which entails the transcription of recordings by five expert annotators who compare them against the standard text in the corpus. Each annotator calculates the percentage of accurately transcribed words, and the average of these five values is used to derive the intelligibility score for each dysarthric speaker. We define the results evaluated by annotation accuracy as subjective intelligibility.

The second criterion is recognition accuracy. We utilize open-source ASR models (Paraformer\cite{paraformer} and WENET\cite{wenet}) to transcribe the speech of each speaker and calculate the Word Error Rate (WER). We refer to the outcomes assessed by recognition accuracy as objective intelligibility.

\section{Description of Dysarthria WWS System}
\begin{figure*}[!t]
  \centering
  \includegraphics[width=\linewidth]{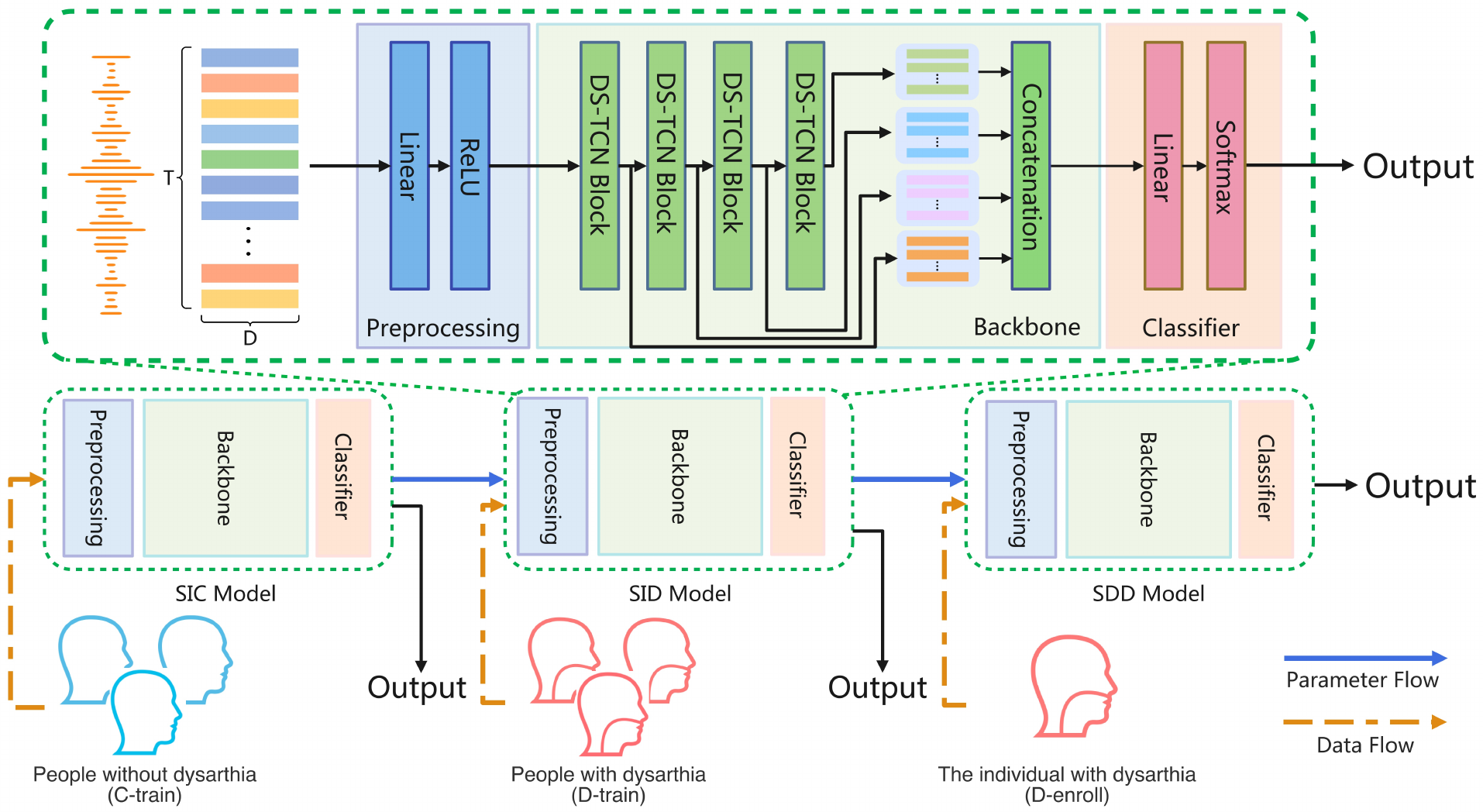}
  \caption{The framework of speaker-dependent dysarthria WWS. The upper part displays the specific network details, where the network architecture of the Speaker-independent Control (SIC), Speaker-independent Dysarthria (SID), and Speaker-dependent Dysarthria (SDD) WWS models are identical. The lower part illustrates the relationships among these models and the overall training process. The SIC model is trained using this network architecture on the C-train dataset. Based on the SIC model, the SID model is fine-tuned with the D-train dataset. Furthermore, the SDD model specific to an individual is fine-tuned using their corresponding D-enroll dataset, building upon the foundation of the SID model.}
  \label{framework}
\end{figure*}

\subsection{Baseline WWS Models}

We utilize the framework provided by the WEKWS toolkit\cite{wekws}. The model architecture of the baseline method consists of four components, starting with a global cepstral mean and variance normalization (CMVN)\cite{cmvn} layer, which normalizes the input acoustic features to follow a Gaussian distribution. Next is a preprocessing module that maps the dimensionality of the input features to the desired dimension. Following that is the backbone network, where we employ a depthwise separable temporal convolutional network (DS-TCN)\cite{dstcn} as the backbone. At the end of the model, as described in\cite{multi}, for each keyword, we add a separate binary classifier after the backbone network to handle scenarios involving multiple keywords. For the MDSC, there are ten keywords, resulting in the network having ten binary classifiers at the final stage.

We employ the following data augmentation techniques inspired by \cite{aug3}:
\begin{itemize}
\item \textbf{Spectrogram augmentation.} We use frequency and time masking for each randomly selected audio.

\item \textbf{Speed perturbation.} The audio speed is randomly changed to be faster or slower, with the ratio in the range [0.9,1.1].

\item \textbf{White noise.} We add white noise to the original audio signals with a random SNR from -15 to 15 dB.
\end{itemize}

We use the training set of Control (C-train) and apply data augmentation techniques before feeding it into the network. The resulting model serves as the Speaker-independent Control (SIC) WWS model.

To accommodate the variability in dysarthric speech, we construct a speaker-independent model that can be generalized across individuals with different speech characteristics. We use the SIC model and conduct fine-tuning using the augmented Dysarthria training set (D-train). The resulting model is called the Speaker-independent Dysarthria (SID) WWS model.

\subsection{Speaker-dependent Dysarthria WWS Model}

Acknowledging the substantial inter-individual variability and limited data volume in dysarthria, we develop a speaker-dependent model to cater to each individual's unique speech characteristics. We employ the SID model for pre-training and perform fine-tuning using augmented personalized enrollment utterances from each individual (D1-enroll to D6-enroll). Enrollment utterances are specific speech samples reserved for each individual in the dataset, typically used in speaker-dependent training or speaker verification tasks~\cite{enroll1,enroll2}. This process yields the Speaker-dependent Dysarthria (SDD) WWS model.

We investigate the impact of the ratio of positive and negative instances and the overall duration of enrollment utterances in the speaker-dependent WWS experiments. To commence, we designate $10$ distinct wake-up words, each with varying content and collectively spanning approximately 30 seconds, as our fixed positive instances. The negative instances are randomly selected from recordings of non-wake-up word utterances by the same individuals, ensuring a proportional duration as required for each session. Subsequently, we gradually vary the number of training samples for both categories to effectively alter the duration of enrollment utterances while maintaining a constant ratio between wake-up and non-wake-up word instances.

\section{Experiments and Analysis}

\subsection{Metrics}
Following the \cite{dku}, the combination of False Reject Rate (FRR) and False Alarm Rate (FAR) is adopted as the criterion, which is defined as follows:
\begin{equation*}
Score = FRR + FAR = \frac{N_{FR}}{N_{wake}}+\frac{N_{FA}}{N_{non-wake}} 
\end{equation*}
where \(N_{wake}\) and \(N_{non-wake}\) denote the number of samples with and without wake-up words in the evaluation set, respectively. \(N_{FR}\) denotes the number of samples containing the wake-up word while not recognized by the system. \(N_{FA}\) denotes the number of samples containing no wake words while predicted to be positive by the system. The lower \textit{Score}, the better the system performance.

\subsection{Analysis of the Baseline WWS Models}
\begin{figure}[!t]
  \centering
  \includegraphics[width=\linewidth]{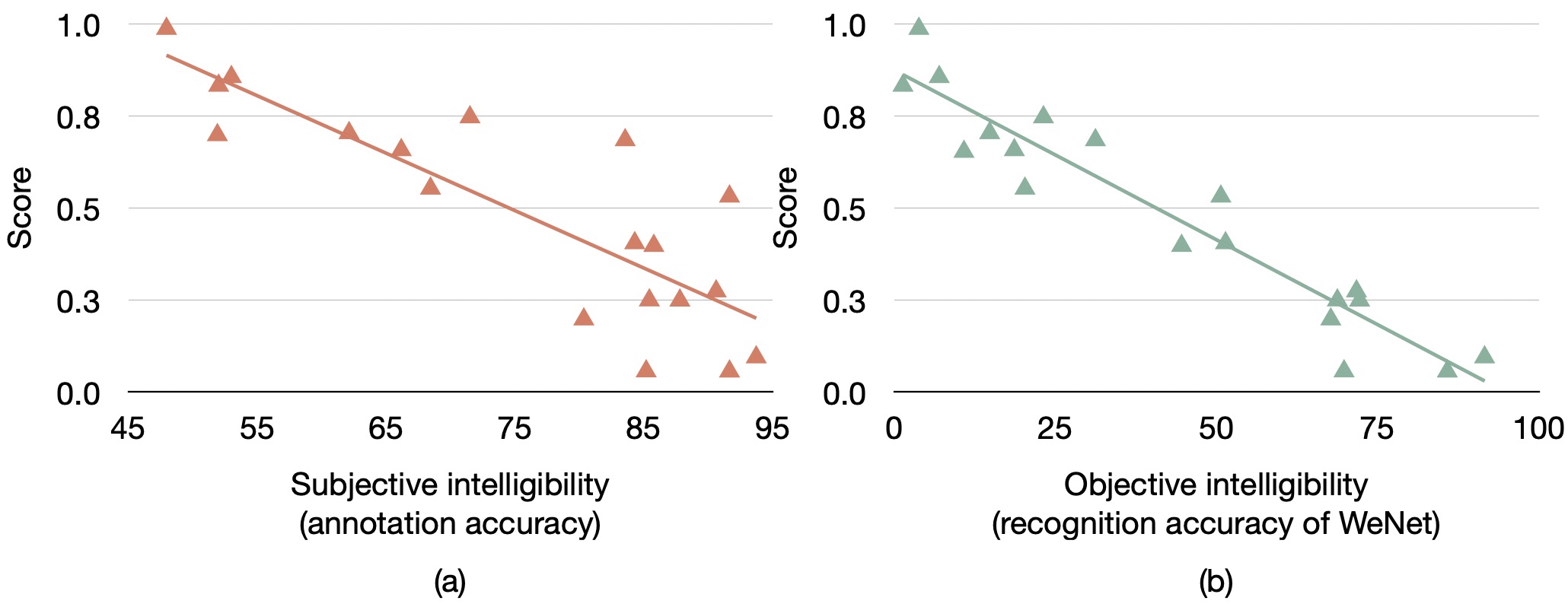}
  \caption{Intelligibility-score relationship for individuals with dysarthria on a conventional WWS system.}
  \label{intell-score}
\end{figure}

Due to the disparities between the wake-up words used in commercial WWS systems and those in the MDSC, it is not feasible to directly test the wake-up performance for dysarthric individuals using commercial systems. Consequently, we test all samples of individuals with dysarthria using the SIC model to examine the performance of the conventional WWS system on dysarthric speech. Figure~\ref{intell-score} is the intelligibility-score relationship graph. We observe that higher intelligibility scores are associated with better WWS performance, indicating a strong positive correlation between these two variables.

\begin{table}[th]
  \caption{Experimental results for SIC, SID and SDD models}
  \setlength{\tabcolsep}{3.2mm}
  \label{baseline}
  \centering
  \begin{tabular}{ c |c| c c c c}
    \toprule
    \multicolumn{1}{c}{\textbf{Model}} &
    \multicolumn{1}{c}{\textbf{Test Set}} &
    \multicolumn{1}{c}{\textbf{FAR$\downarrow$}} &
    \multicolumn{1}{c}{\textbf{FRR$\downarrow$}} &
    \multicolumn{1}{c}{\textbf{Score$\downarrow$}} \\
    \midrule
    \textbf{SIC} & \textbf{C-test} & 0.0148 & 0.0000 & 0.0148   \\
    \textbf{SIC} & \textbf{D-test} & 0.1630 & 0.3708 & 0.5339   \\
    \textbf{SID} & \textbf{D-test} & 0.1538 & 0.1875 & 0.3413   \\
    \textbf{SDD} & \textbf{D-test} & 0.0555 & 0.0833 & 0.1388 \\
    \bottomrule
  \end{tabular}
\end{table}

Moreover, we conduct separate tests on the SIC model using the C-test and D-test to evaluate the distinctions between speech samples of individuals with and without dysarthria. We also test the SID model on the D-test. From the results of the SIC models in Table~\ref{baseline}, we observe a performance discrepancy of the SIC model between the C-test and D-test, indicating significant speech feature differences between individuals with and without dysarthria. It is challenging to create a system applicable to dysarthric individuals solely based on existing datasets from non-dysarthric people. Additionally, comparing the experimental results of SIC and SID on the D-test, we discover a notable improvement when incorporating dysarthric speech data.

\begin{figure}[!t]
  \centering
  \includegraphics[width=\linewidth]{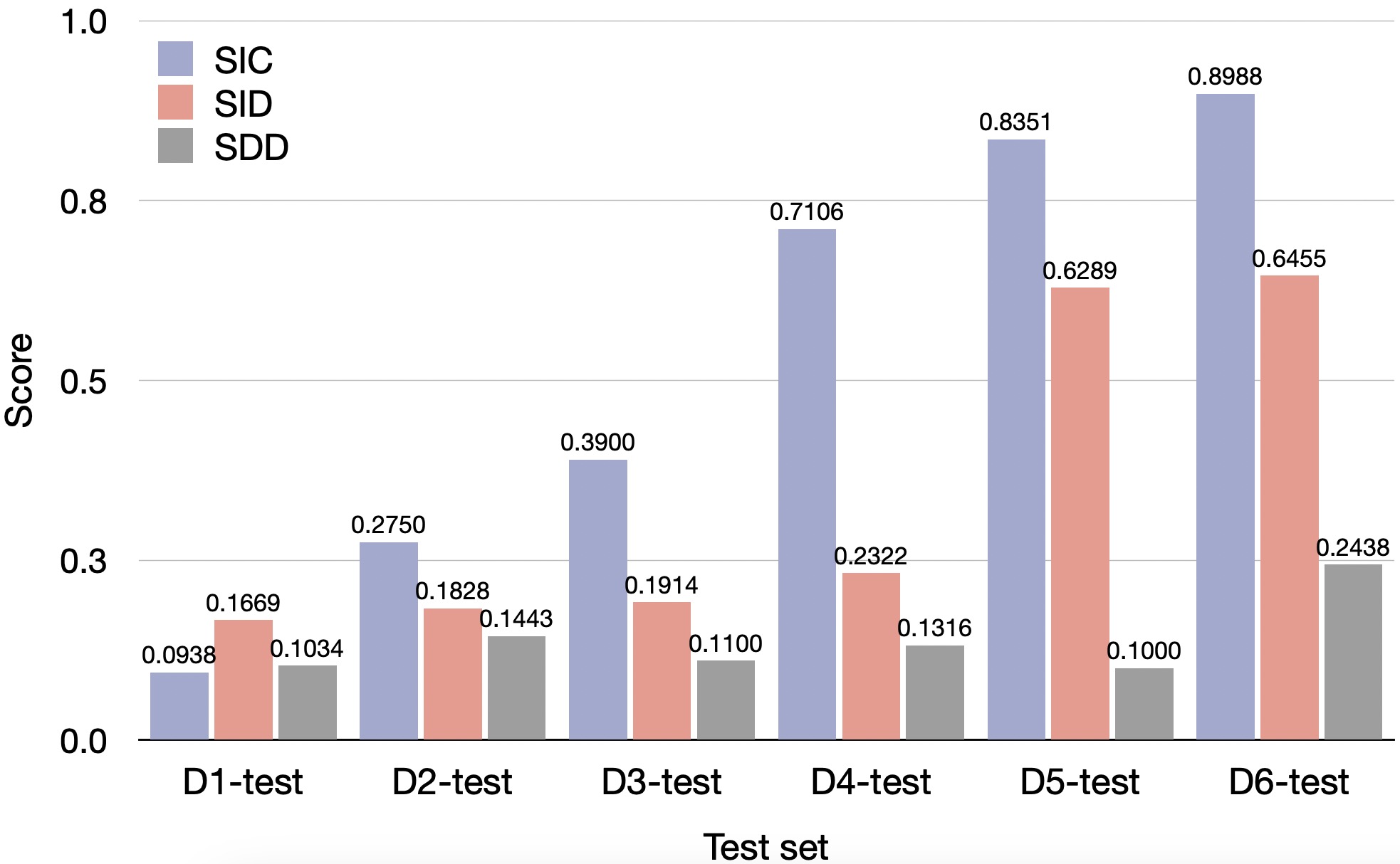}
  \caption{Wake-up performance of SIC, SID and SDD models on D1-D6 test sets. A lower score indicates better performance.}
  \label{score}
\end{figure}

Furthermore, as shown for SIC and SID in Figure~\ref{score}, we provide the testing results for each individual (D1-D6) within the D-test subset using the SIC and SID models to assess the impact of intelligibility on model enhancement. It is apparent that individuals with moderate intelligibility, namely D3 and D4, demonstrate the most substantial relative enhancements, achieving improvements of $51\%$ and $67\%$, respectively. This may be attributed to D3 and D4 representing the average speech features of dysarthria, striking a balance between complexity and simplicity and thereby making them more representative of a wider population. However, the improvement is not general; note that for the individual with the highest intelligibility, D1, the results worsened after fine-tuning with D-train. This unexpected result can be attributed to D1's pronunciation characteristics being more different from those of the other dysarthric individuals. Therefore, fine-tuning the model on the dysarthria dataset results in a decrease in performance.

Based on the above analysis, we can summarize several distinct characteristics of dysarthric speech:
\begin{itemize}
\item \textbf{Significant in-domain variance.} Each dysarthric individual exhibits unique speech characteristics, leading to substantial variations in pitch, speech rate, breath patterns, and sentence boundaries.
\item \textbf{Limited data volume.} It is reflected in the scarcity of recruiting dysarthric individuals and the difficulty of recording. Due to the disorder's impact, it is often challenging to conduct prolonged recording sessions.
\end{itemize}

Therefore, we propose a speaker-dependent dysarthria WWS model as a promising direction. Adapting the system for each speaker can ignore in-domain variance and requires only a tiny amount of speaker-specific data.

\subsection{Analysis of Speaker-dependent Dysarthria WWS Model}

\begin{figure}[t]
  \centering
  \includegraphics[width=\linewidth]{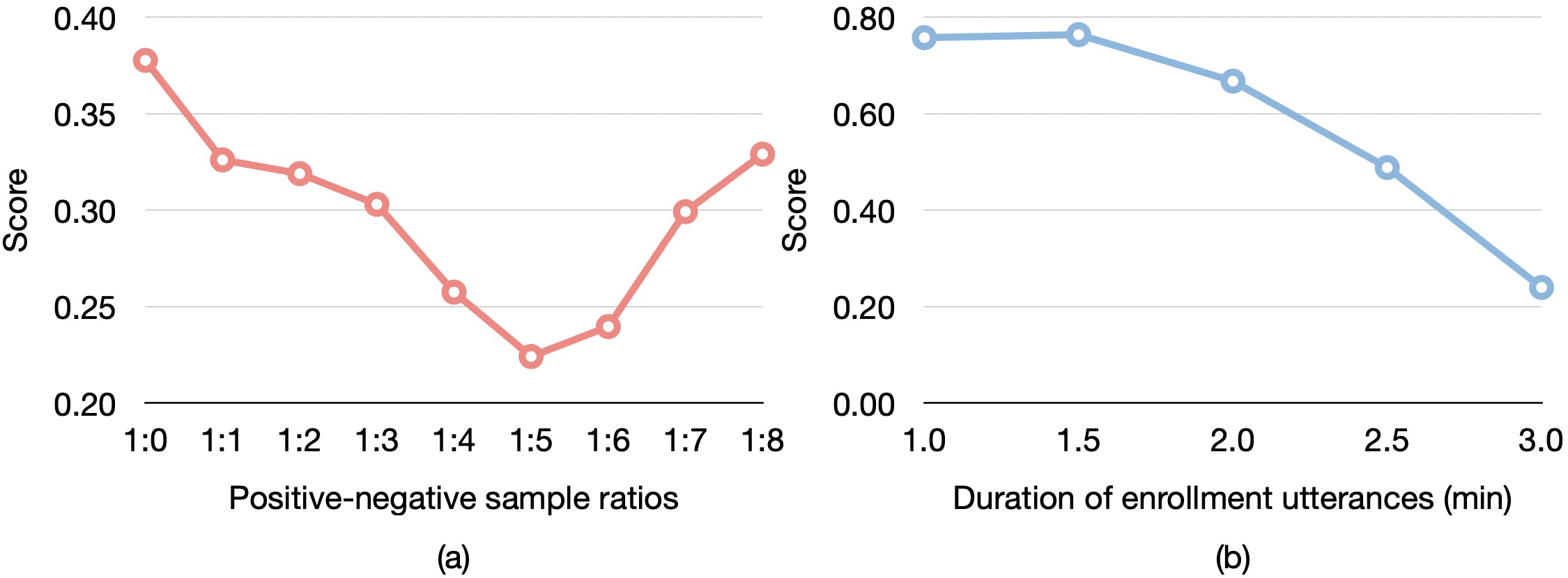}
  \caption{(a) The performance for different positive-to-negative ratios of enrollment utterances. The x-axis represents the duration ratio of positive to negative instances, ranging from 1:0 to 1:10. (b) The performance for different durations of enrollment utterances. The x-axis represents the entire duration of enrollment utterances, ranging from 1 minute to 3 minutes.}
  \label{enrollment}
\end{figure}

To investigate the wake-up performance of the positive-to-negative sample ratios and the duration of enrollment utterances, we randomly select D2 as the research subject. The results are depicted in Figure~\ref{enrollment}. Considering all factors, we utilize enrollment utterances with a total duration of 3 minutes and a wake-up to non-wake-up word ratio of 1:5 to strike a balance between the duration and performance.

After that, we conduct tests on D1 to D6 test sets using the corresponding SDD model and record their average results in the SDD model of Table~\ref{baseline}. Utilizing only 3-minute enrollment utterances for model customization leads to significant improvements. This demonstrates the potential application of speaker-dependent approaches. Moreover, we also give the results of the SDD model tested on D1-D6 in Figure~\ref{score}. By comparing it with the SIC and SID models, it becomes apparent that the wake-up performance of the SDD model is minimally affected by intelligibility, indicating its robustness in handling various levels of intelligibility. However, despite the notable improvement, individuals with extremely low intelligibility (D6) still exhibit relatively poor results. This highlights the ongoing challenge in cases of severe dysarthria. Further investigation and adaptation techniques are required to address the specific needs of individuals with extremely low intelligibility.

\begin{figure}[t]
  \centering
  \includegraphics[width=\linewidth]{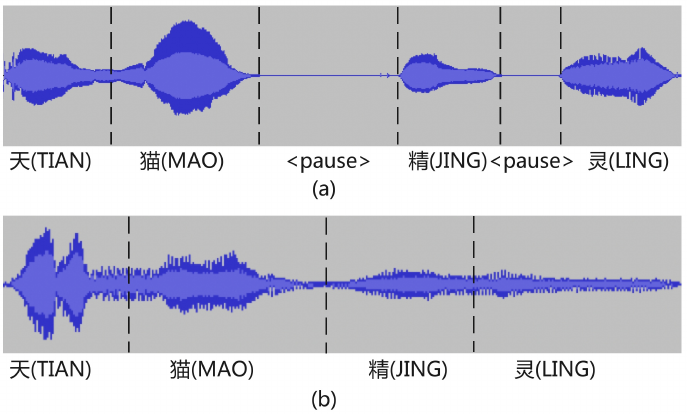}
  \caption{The wake-up failure cases of individuals with dysarthria. In case (a), the speaker exhibits sudden pauses and breaths within the sentence. In case (b), the speaker experiences a decreased volume in the latter half of the phrase, accompanied by a noticeable prolongation of sounds and instances of speech coarticulation.}
  \label{failure}
\end{figure}

As shown in Figure~\ref{failure}, we also observe failure cases. These samples demonstrate unique speech characteristics specific to individuals with dysarthria, which may challenge the system to make accurate judgments.

\section{Conclusion}
In this paper, firstly, we construct and release the Mandarin Dysarthria Speech Corpus (MDSC), providing a valuable resource for dysarthria wake-up research in home environments. Secondly, we conduct a comprehensive experimental analysis on MDSC, introducing realistic challenges for dysarthria wake-up systems. Finally, we propose a customized dysarthria WWS system that demonstrates robustness in handling intelligibility and achieving exceptional performance. This study represents an initial exploration of dysarthric speech. We intend to delve further into language-related research for dysarthria. By doing so, our goal is to foster greater societal awareness and comprehension of dysarthria while actively working towards eradicating discrimination and prejudice against individuals with this condition.

\section{Acknowledgements}
This work was supported by the National Natural Science Foundation of China under Grant No.62171427.

\bibliographystyle{IEEEtran}
\bibliography{mybib}

\end{document}